\documentclass[sigconf,nonacm]{acmart}
\usepackage[english]{babel}
\usepackage{colortbl}
\usepackage{subcaption}
\usepackage{tikz}
\usepackage{placeins}
\usepackage{enumitem}
\usepackage{amsmath,booktabs,multirow,makecell,arydshln}

\title{Probe Before You Learn: Allocating Prompts from Initial Gradients under Label Noise}
\setcopyright{none}
\author{Jiazhe Zhang}
\affiliation{\institution{Institute of Computing Technology, Chinese Academy of Sciences}\institution{University of Chinese Academy of Sciences}\city{Beijing}\country{China}}

\author{Zibin Geng}
\affiliation{\institution{Institute of Computing Technology, Chinese Academy of Sciences}\institution{University of Chinese Academy of Sciences}\city{Beijing}\country{China}}

\author{Qiang Ma}
\affiliation{\institution{Qiyuan Lab}\city{Beijing}\country{China}}

\author{Xuefeng Jiang}
\affiliation{\institution{Institute of Computing Technology, Chinese Academy of Sciences}\institution{University of Chinese Academy of Sciences}\city{Beijing}\country{China}}

\author{Yijun Lin}
\affiliation{\institution{People's Public Security University of China}\city{Beijing}\country{China}}

\author{Jingyi Wang}
\affiliation{\institution{Qiyuan Lab}\city{Beijing}\country{China}}

\author{Yu Yan}
\affiliation{\institution{Institute of Computing Technology, Chinese Academy of Sciences}\institution{University of Chinese Academy of Sciences}\city{Beijing}\country{China}}

\author{Xingzhou Zhang}
\affiliation{\institution{Institute of Computing Technology, Chinese Academy of Sciences}\institution{University of Chinese Academy of Sciences}\city{Beijing}\country{China}}

\author{Min Liu}
\affiliation{\institution{Institute of Computing Technology, Chinese Academy of Sciences}\institution{University of Chinese Academy of Sciences}\city{Beijing}\country{China}}

\author{Yuwei Wang}
\affiliation{\institution{Institute of Computing Technology, Chinese Academy of Sciences}\city{Beijing}\country{China}}

\author{Sheng Sun}
\authornote{Corresponding author.}
\affiliation{\institution{Institute of Computing Technology, Chinese Academy of Sciences}\city{Beijing}\country{China}}
\email{sunsheng@ict.ac.cn}
\renewcommand{\shortauthors}{Zhang et al.}

\begin{document}
\begin{abstract}
Prompt learning provides a parameter-efficient way to adapt pretrained vision--language models through text and visual prompts, but noisy labels prevalent in real-world data inevitably bias prompt updates and thereby undermine adaptation effectiveness. 
Prior work on robust prompt learning focuses on suppressing unreliable updates under noisy supervision. 
We uncover a pivotal yet overlooked aspect--\textbf{prompt allocation}: shifting the parameter budget between text and visual prompts can substantially enhance robustness while keeping the training objective and total budget fixed.
However, suitable prompt allocations vary across tasks and are unknown a priori, rendering brute-force candidate training computationally prohibitive.
We propose GPrompt, a training-free method that derives allocation between text and visual prompts using only loss gradients at initialization before prompt optimization.
Using a small sample of images and generated pseudo-labels as shared supervision, GPrompt performs an initialization probe to extract paired text and visual loss gradients, and characterizes the geometry of their trace-normalized Gram matrices through both concentration within each modality and overlap between modalities.
GPrompt then derives a closed-form allocation, and projects it onto feasible prompt counts under the exact parameter budget, enabling task-specific selection without training candidate configurations.
Experiments across eleven datasets show that GPrompt outperforms the strongest baseline by up to 7.75\% while selecting configurations close to the best allocation identified through exhaustive training. 
The average runtime of gradient probe across datasets is 99.9\% lower than that of exhaustive candidate training.
\end{abstract}

\maketitle
\begin{figure}[t]
\centering
\includegraphics[width=\columnwidth]{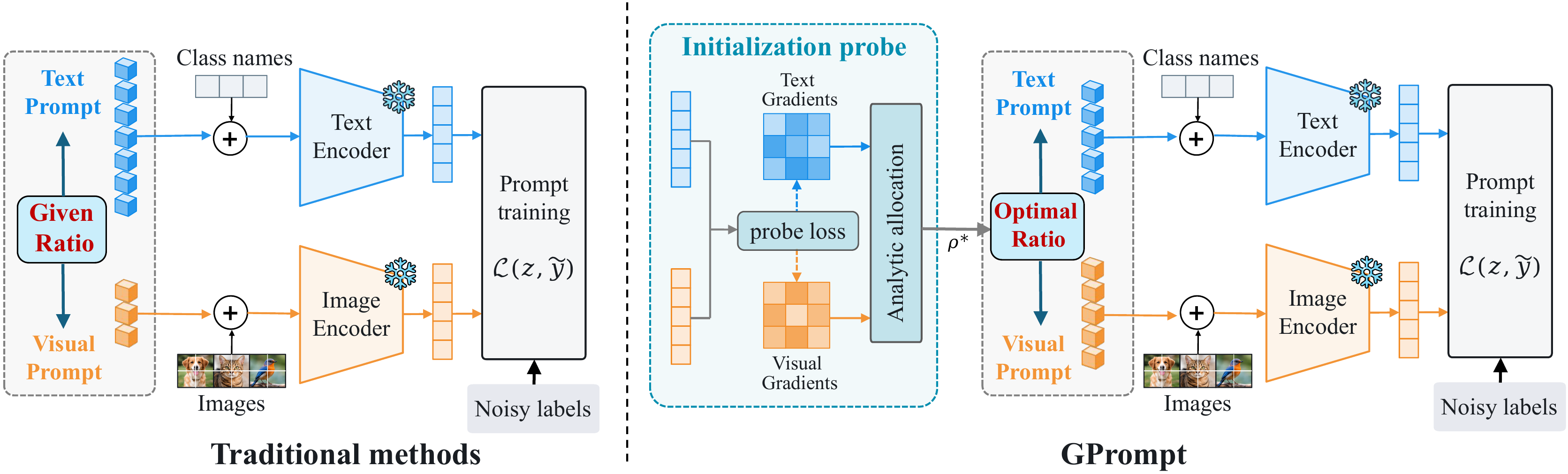}
\caption{Prompt allocation before training. Traditional methods use a preset allocation between text and visual prompts (left); GPrompt selects the allocation from paired gradients at initialization before prompt training (right). Both configurations use the same prompt parameter budget, with cubes indicating the budget assigned to each modality. Both encoders remain frozen, and observed noisy labels are used to supervise prompt training.}
\Description{The left panel shows text and visual prompts with a given allocation. They are combined with class names and image patches, respectively, and passed through frozen encoders for prompt training with noisy labels. The right panel adds an initialization probe: a shared probe loss yields paired text and visual gradients, which determine an analytic allocation before prompt training. Snowflakes mark frozen encoders, and blue and orange cubes represent text and visual prompt allocations.}
\label{fig:teaser}
\end{figure}

\section{Introduction}
\label{sec:intro}

Pretrained vision--language models such as CLIP align visual content with natural language captions in a shared embedding space~\citep{CLIP2021}, enabling strong zero-shot recognition and effective transfer to diverse downstream tasks~\citep{ALIGN2021,LiT2022,SigLIP2023}. 
Prompt learning is a parameter-efficient way to adapt these models by optimizing a small set of prompt vectors for each task while keeping the pretrained backbone frozen~\citep{CoOp2022,VPT2022}. 
Textual and visual prompts modulate the shared embedding space through separate pathways: Text prompt vectors in the text encoder reshape class prototypes into task-specific representations~\citep{CoOp2022,CoCoOp2022,ProDA2022}, whereas visual prompts are inserted alongside image patch embeddings into the image encoder to steer visual patterns~\citep{VPT2022,FacingVPT2024}. Jointly learning textual and visual prompts improves image--text alignment and provides complementary pathways for downstream adaptation~\citep{MaPLe2023,DPT2024}.

However, label noise is inevitable in large-scale image data collected from the web, and mislabeled samples generate misleading loss gradients that bias prompt optimization~\citep{Clothing2015,WebVision2017,NoiseSurvey2023}.
These erroneous updates contaminate the shared prompt parameters, allowing unreliable supervision to propagate across samples and potentially distort cross-modal alignment~\citep{PTNL2023,NLPrompt2025}.
Following paradigms for learning with noisy labels, including loss correction, sample selection, and regularization against memorization~\citep{LossCorrection2017,CoTeaching2018,ELR2020}, existing approaches to robust prompt learning primarily suppress unreliable gradient updates to prevent performance degradation. 
Sample selection and label refinement identify trustworthy examples or correct their targets~\citep{JoAPR2024,DeFT2024,DKAF2026}, whereas robust objectives reduce the contributions of corrupted gradients during optimization~\citep{NLPrompt2025,DSPT2026}.

Existing robust prompt learning methods either optimize a single prompt modality or retain a fixed configuration of textual and visual prompts, leaving modality-specific susceptibility to label noise unexplored.
Since text and visual prompts modulate logits through distinct encoder Jacobians, the identical noisy supervision induces divergent gradient geometries varying across downstream tasks.
Under a fixed trainable-parameter budget, allocation between text and visual prompts determines each pathway’s adaptation capacity and thus the robustness of joint prompt learning.
An unsuitable split can amplify the influence of concentrated, noise-contaminated gradients in one modality while limiting complementary directions from the other, thus degrading cross-modal alignment and robust generalization.
This motivates a pivotal question to resolve: \textbf{how should the available trainable parameter budget be allocated between textual and visual prompts under label noise?}

We first investigate this question through \textbf{experimental observation}. 
We vary the text-to-visual parameter ratio while holding the robust objective, total trainable parameter budget, data splits, and training setting fixed. \textit{The key observation is that this allocation ratio markedly influences classification accuracy, and the optimal ratio varies substantially across datasets (Figure~\ref{fig:allocation_curves}).} 
For example, under asymmetric noise, Oxford Flowers performs best at a ratio of 4:6, whereas the optimum on UCF101 is 7:3. These results show that no fixed allocation generalizes across tasks, motivating task-adaptive allocation before prompt optimization.

Next, \textbf{modality-gradient analysis} establishes a principled basis for prompt allocation.
We characterize the geometry of paired textual and visual gradients, and introduce two spectral signatures: 
\textit{concentration within each modality}, which indicates whether prompt gradients collapse onto a few dominant directions, and \textit{overlap between modalities}, which quantifies redundancy or complementarity between textual and visual profiles. 
Together, these metrics define a gradient-guided allocation criterion that minimizes the spectral concentration of joint gradient profile.

Building on this analysis, we propose \textbf{GPrompt}, a training-free method that determines task-specific allocation between textual and visual prompts before optimization.
GPrompt proceeds in three stages: common gradient probing, modality gradient profiling, and analytic budget allocation.
The initial probe extracts paired gradients from a small image sample using pseudo-labels at initialization, without relying on noisy supervision.
The gradient profiling stage uses trace-normalized Gram matrices to quantify concentration within each modality and overlap between modalities.
The analytic allocator derives a closed-form budget allocation by minimizing the spectral concentration of joint gradient profile, which is projected onto feasible prompt counts satisfying budget constraint.
Only the selected prompt configuration is subsequently trained with both encoders frozen.
Across eleven image classification datasets under symmetric, asymmetric, and natural label noise, GPrompt outperforms the strongest baseline by up to 7.75\% and achieves mean accuracy within 0.131\% of the best allocation identified through exhaustive candidate training under asymmetric noise.

Our contributions are summarized as follows:
\begin{itemize}
\item We first investigate robust prompt optimization from the previously unexplored perspective of textual and visual prompt allocation, revealing that modality-specific sensitivities to noise label necessitates task-adaptive allocation for robustness improvement.
\item We introduce GPrompt, a training-free allocation framework that determines prompt allocation at initialization through paired text--visual gradient profiling and projection onto feasible prompt counts under the exact budget constraint, avoiding exhaustive candidate training.
\item Across eleven datasets under diverse noise conditions, GPrompt outperforms the state-of-the-art by up to 7.75\% while selecting near-optimal configuration, and the allocation assessment time of GPrompt is 99.9\% lower than that of exhaustive candidate training.
\end{itemize}

\begin{figure*}[t]
\centering
\begingroup
\setlength{\unitlength}{\dimexpr\textwidth/2172\relax}
\begin{picture}(2172,521)
\put(0,0){\includegraphics[width=\textwidth]{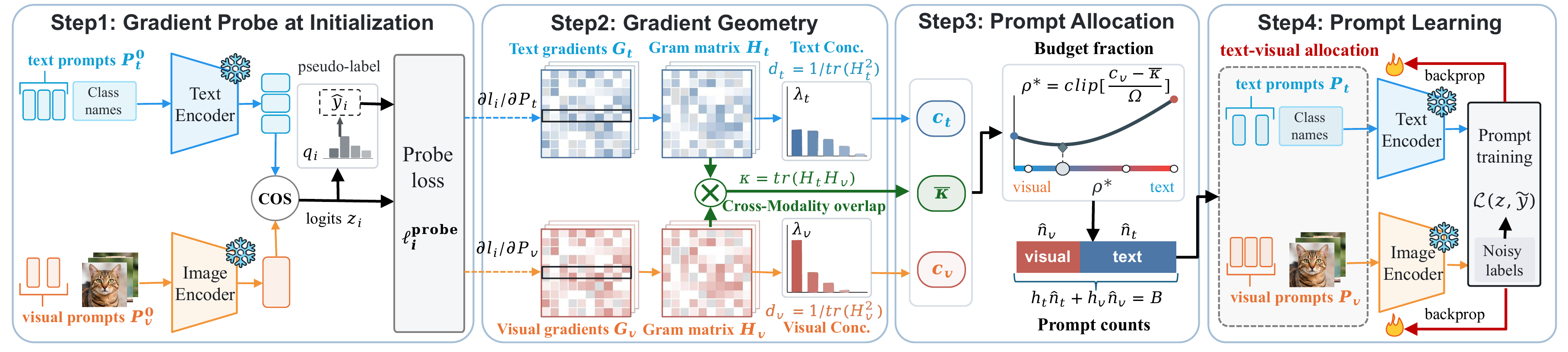}}
\put(1165,129){\color[RGB]{248,250,251}\rule{51\unitlength}{32\unitlength}}
\put(1168,138){\color[RGB]{87,87,87}\fontsize{7}{7}\selectfont$1/d_v$}
\end{picture}
\endgroup
\caption{Overview of GPrompt. (1)~Gradient probing at initialization with pseudo-labels, without parameter updates. (2)~Gradient geometry: effective dimensions $d_t$, $d_v$ and overlap $\kappa$. (3)~Closed-form budget allocation: the optimal text parameter fraction $\rho^*$ is mapped to feasible prompt counts $(\hat n_t,\hat n_v)$. (4)~Training of the selected prompts with the DSPT loss on noisy labels.}
\Description{Four panels from left to right. Panel 1: initialized text and visual prompts pass through the frozen text and image encoders; the logits of the sampled images give pseudo-labels and a shared loss, and no parameter is updated. Panel 2: per-image gradients of the two branches form normalized Gram matrices with their eigenvalue spectra, effective dimensions, and overlap. Panel 3: the averaged coefficients define a convex quadratic in the text fraction whose minimizer is mapped to the nearest feasible token numbers under the budget. Panel 4: only the selected text and visual prompts are trained with the DSPT loss on noisy labels.}
\label{fig:two_stage_model}
\end{figure*}

\section{Related Work}
\label{sec:related}

\textbf{Prompt learning for vision--language models.}
CLIP classifies images by comparing image representations with text embeddings of class descriptions~\citep{CLIP2021}. CoOp learns continuous context vectors that the text encoder processes with class names~\citep{CoOp2022}, and CoCoOp conditions these vectors on the input image~\citep{CoCoOp2022}. Visual prompt tuning inserts learnable tokens into a frozen vision transformer~\citep{VPT2022}. 
MaPLe couples prompts in the vision and language branches~\citep{MaPLe2023}. 
KgCoOp~\citep{KgCoOp2023}, ProGrad~\citep{ProGrad2023}, and PromptSRC~\citep{PromptSRC2023} preserve pretrained knowledge through feature- or gradient-based regularization during adaptation.
CLIP-Adapter blends adapted features with pretrained representations~\citep{CLIPAdapter2024}, and Tip-Adapter constructs a cache from labeled examples for few-shot classification~\citep{TipAdapter2022}. 
Although these methods provide complementary adaptation pathways, joint prompting typically adopts preset text and visual capacities. GPrompt instead determines their allocation under a fixed parameter budget and noisy supervision.

\noindent
\textbf{Learning with noisy labels.}
Existing methods for learning with noisy labels include loss correction~\citep{LossCorrection2017}, reliable sample selection~\citep{CoTeaching2018, DivideMix2020}, regularization against memorization~\citep{ELR2020}, and robust objective design.
Representative robust objectives include mean absolute error (MAE)~\citep{RobustLoss2017}, generalized cross entropy~\citep{GCE2018}, normalized losses with complementary terms~\citep{APL2020}, and prediction clipping~\citep{LogitClip2023}.
Sharpness-aware minimization~\citep{SAM2021} also improves robustness to label noise by regulating example gradients~\citep{SAMNoise2024,SANER2026}. 
For vision--language adaptation, prompt tuning exhibits inherent resilience to label noise~\citep{PTNL2023}. JoAPR partitions samples and refines labels~\citep{JoAPR2024}; DeFT identifies noisy labels using positive and negative text prompts~\citep{DeFT2024}; and DKAF combines multimodal pseudo-labels with label refinement~\citep{DKAF2026}. 
NLPrompt employs optimal transport for sample purification, and applies cross entropy to the clean subset and MAE to the noisy subset~\citep{NLPrompt2025}. 
DSPT suppresses high-error examples through two successive softmax operations~\citep{DSPT2026}.
NA-MVP aligns image patches with multiple prompt views for few-shot learning with noisy labels~\citep{NAMVP2026}. 
These robust prompt learning methods improve robustness under preset prompt capacities, while GPrompt pioneers optimization robust to label noise through allocation between text and visual prompts according to gradient geometry of the two modalities.

\noindent
\textbf{Evaluation metrics and gradient analysis.}
LogME estimates model transferability from extracted features and target labels~\citep{LogME2021}. 
RankMe assesses self-supervised representations using effective rank without labels~\citep{RankMe2023}.
GradSign uses gradients for each sample at initialization for architecture assessment~\citep{GradSign2022}, and InTrain combines participation ratios of the activation covariance matrices with cumulative gradient health ~\citep{InTrain2026}.
Gradient inner products further characterize training dynamics and sample interactions, as exemplified by neural tangent kernels~\citep{NTK2018}.
TracIn estimates the influence of training examples from loss gradients at checkpoints~\citep{TracIn2020}.
Unlike the above methods, GPrompt constructs paired Gram matrices of text and visual loss gradients in a shared space of per-image losses, where concentration within each modality and overlap between modalities are quantified to derive prompt allocation for each task.

\section{Method}
\label{sec:method}

GPrompt allocates a fixed parameter budget between text and visual prompts before training to improve robustness under label noise. A common gradient probe extracts paired loss gradients at initialization (Section~\ref{sec:probe}). Gradient profiling then measures concentration within each modality and overlap between modalities in a shared space of image losses (Section~\ref{sec:geometry}). These quantities yield a closed-form allocation, which is projected onto feasible prompt counts under the exact budget constraint (Section~\ref{sec:selection}). Figure~\ref{fig:two_stage_model} illustrates the pipeline.

\subsection{Problem formulation}
\label{sec:problem}
\label{sec:gap}
We adapt a pretrained CLIP model~\citep{CLIP2021} to a $C$-class target dataset $\mathcal D=\{(x_i,\tilde y_i)\}_{i=1}^{N}$, where $\tilde y_i$ is the potentially corrupted observed label of image $x_i$. 
The text encoder $f_t$ and image encoder $f_v$ remain frozen. 
The text prompt matrix $P_t\in\mathbb R^{n_t\times h_t}$ comprises $n_t$ learnable vectors concatenated with each tokenized class name, and the visual prompt matrix $P_v\in\mathbb R^{n_v\times h_v}$ comprises $n_v$ learnable tokens inserted into patch embeddings of each image~\citep{VPT2022}, where $h_t$ and $h_v$ are the token embedding dimensions of the text and image encoders, respectively. 
Let $\theta_t=\operatorname{vec}(P_t)$, $\theta_v=\operatorname{vec}(P_v)$, and $\theta=(\theta_t,\theta_v)$.
For the training image $x_i$, let $z_i=(z_{i,1},\ldots,z_{i,C})$ denote the logit vector of $x_i$, where $z_{i,c}$ is its class-$c$ logit.
$z_{i,c}(x_i;\theta)=\exp(s)\,\operatorname{cos}\bigl(f_v(x_i;P_v),f_t(c;P_t)\bigr)$, where $f_t$ and $f_v$ denote the prompted class and image embeddings, respectively. $s$ is the logit-scale parameter learned during CLIP pretraining and kept frozen.
We optimize prompts using the double-softmax loss~\citep{DSPT2026}. The loss of $x_i$ is $\ell(z_i,\tilde y_i)=-\log[\operatorname{softmax}(\operatorname{softmax}(z_i))]_{\tilde y_i}$.

An allocation $a=(n_t,n_v)$ specifies the numbers of text and visual prompt tokens. For a fixed trainable parameter budget $B>0$, we consider a nonempty set of feasible allocations:
\begin{equation}
\mathcal A_B\subseteq\{(n_t,n_v)\in\mathbb Z_{\geq0}^2:\,
n_t\times h_t+n_v\times h_v=B\}.
\label{eq:budget}
\end{equation}
For an allocation $a$, the text parameter fraction is $\rho(a)=h_tn_t/B$, i.e., the fraction of the total prompt parameter budget allocated to text prompts, and the visual parameter fraction is $1-\rho(a)$. 
GPrompt seeks to select an allocation $a\in\mathcal A_B$ before prompt learning without training all candidate configurations.

\subsection{Paired gradient probe}
\label{sec:probe}
\noindentparagraph{\normalfont\bfseries Pseudo-label generation.}
We sample $M$ training images and initialize $n_t^{\mathrm p}$ text and $n_v^{\mathrm p}$ visual prompt tokens at the probe stage. 
Let $\theta^0=(\theta_t^0, \theta_v^0)$ denote the joint initial prompt parameters, where $D_t=n_t^{\mathrm p}\times h_t$ and $D_v=n_v^{\mathrm p}\times h_v$ are the dimensions of text and visual prompt parameters, respectively.
The probe token counts $(n_t^{\mathrm p},n_v^{\mathrm p})$ are fixed across tasks. 
Both prompt types use the same pseudo-label for each image. We first average the initial predicted class probabilities for the original image and its horizontal flip:
\begin{equation}
q_i=\tfrac12\bigl[\operatorname{softmax}(z(x_i;\theta^0))+\operatorname{softmax}(z(\operatorname{flip}(x_i);\theta^0))\bigr].
\label{eq:pseudo}
\end{equation}
The pseudo-label is $\hat y_i=\arg\max_c q_{i,c}$. 
For each image, both text and visual prompt gradients are computed with the fixed pseudo-label. The observed labels $\tilde y_i$ are used only for prompt training after the allocation between text and visual prompts is selected.

\noindentparagraph{\normalfont\bfseries Gradients for each image.}
We use the same double-softmax form for $\ell_i^{\mathrm{probe}}$, evaluated on the original image with $\hat y_i$ held fixed.
For each image $i$, we evaluate the paired text and visual gradients $(g_i^t, g_i^v)$ with respect to the prompt parameters $\theta^0$ at initialization:
\begin{equation}
g_i^t=\left.\nabla_{\theta_t^0}\ell_i^{\mathrm{probe}}(z(x_i; \theta), \hat y_i)
\right|_{\theta^0}, 
\quad g_i^v=\left.\nabla_{\theta_v^0}\ell_i^{\mathrm{probe}}(z(x_i; \theta), \hat y_i)
\right|_{\theta^0}.
\label{eq:probegrad}
\end{equation}
A single backward pass on the loss $\ell_i^{\mathrm{probe}}$ for each image yields both prompt gradients without updating any parameter. 
Let $r_i=\nabla_{z_i}\ell_i^{\mathrm{probe}}$ denote the probed loss gradient with respect to logits shared by both types of prompts. For $m\in\{t,v\}$, $J_i^m=\left.\frac{\partial z(x_i;\theta)}{\partial\theta_m}\right|_{\theta=\theta^0}$ denotes the Jacobian of the same joint logits with respect to that modality's prompt parameters at initialization.

For any two samples $i$ and $j$, the inner product of the text prompt gradients is $\langle g_i^t,g_j^t\rangle
=r_i^\top J_i^t(J_j^t)^\top r_j$.
For a fixed sample pair, the difference between the two modalities' gradient inner products arises from their respective Jacobian products. The matrix $J_i^t(J_j^t)^\top$ is the $(i,j)$ block of the empirical neural tangent kernel~\citep{NTK2018}.
Similarly, for the visual prompt, we have $\langle g_i^v,g_j^v\rangle
=r_i^\top J_i^v(J_j^v)^\top r_j$.
The paired gradients thus provide a comparison of the two prompt parameterizations under the same supervision.

\subsection{Gradient profiling for each modality}
\label{sec:geometry}
\noindentparagraph{\normalfont\bfseries Gram geometry of per-image gradients.}
For each modality $m\in\{t, v\}$, we stack the per-image gradients row-wise into $G_m\in\mathbb R^{M\times D_m}$ whose $i$-th row is $(g_i^m)^\top$. 
The Gram matrix of per-image gradients is $K_m=G_mG_m^\top$.
Let $\boldsymbol\ell\in\mathbb R^M$ denote the vector of per-image probe losses.
For the vector of sample weights $w\in\mathbb R^M$, a hypothetical gradient-descent step on the weighted probe loss $w^\top\boldsymbol\ell$ updates the modality-$m$ prompt parameters by 
$\delta\theta_m^0=-\eta G_m^\top w$, where $\eta>0$. 
A first-order Taylor expansion at initialization yields
$\Delta\boldsymbol\ell=-\eta K_mw+O(\eta^2).$
This expansion assumes locally smooth losses and holds the other prompt type fixed. 
For a gradient step on image $j$ alone, the first change in the loss of image $i$ is $-\eta (K_m)_{ij}$~\citep{TracIn2020}. 
A positive $(K_m)_{ij}$ indicates that the update from image $j$ decreases the loss of image $i$, whereas a negative value indicates an increase.
For an eigenvector $w$ of $K_m$ with eigenvalue $\lambda$,  weighting the gradient update by $w$ yields $\Delta\boldsymbol{\ell} \approx -\eta \lambda w$ to first order.

\noindentparagraph{\normalfont\bfseries Normalized modality profiling.}
To compare the spectral geometry of the gradients of each modality independently of their magnitudes, we normalize the gradient Gram matrix $K_m$ by its trace, i.e., $T_m=\operatorname{tr}(K_m) =\sum_i\|g_i^m\|^2$ for $K_m\neq 0$, and define $H_m=K_m/T_m$ such that $\operatorname{tr}(H_m)=1$.
Its effective dimension $d_m=\frac1{\operatorname{tr}(H_m^2)}$, also known as the participation ratio of $H_m$, equals $k$ when $K_m$ has $k$ equal nonzero eigenvalues.
It decreases as the normalized eigenvalue distribution becomes more concentrated.
Thus, a larger $d_m$ indicates lower spectral concentration (Appendix~\ref{app:shared_update_analysis}). 
Trace normalization makes $d_m$ invariant to a uniform rescaling of all gradients within one modality, so $d_m$ captures spectral concentration rather than the overall gradient magnitude. 
For gain ranking, we compute $s_{\mathrm{gain}}=\log(d_t/d_v)$ for each paired probe and average the resulting scores. 
The score is used to rank gains from adding visual prompts, while the allocation objective below selects a parameter fraction using both modality concentrations and their overlap.

\noindentparagraph{\normalfont\bfseries Joint gradient profiling under a given allocation.}
To characterize the gradient geometry associated with different allocations of the budget between text and visual prompts using a fixed probe configuration, we use the fixed probe profiles as surrogates for the normalized gradient profiles across candidate parameter counts.
For a candidate allocation with text parameter fraction $\rho\in[0,1]$, we define the joint gradient profile using the parameter fractions as weights:
\begin{equation}
H(\rho)=\rho H_t+(1-\rho)H_v.
\label{eq:joint_response}
\end{equation}
Because $H_t$ and $H_v$ are positive semidefinite with unit trace, $H(\rho)$ is also positive semidefinite and satisfies $\operatorname{tr}(H(\rho))=1$.
Appendix~\ref{app:shared_update_analysis} constructs this approximation through a budget--weighted mixture of the distributions for the two modalities over normalized gradient directions.

We quantify the spectral concentration of the joint gradient profile by
\begin{equation}
F(\rho)=\operatorname{tr}(H(\rho)^2)
=\frac{\rho^2}{d_t}+\frac{(1-\rho)^2}{d_v}
+2\rho(1-\rho)\kappa.
\label{eq:joint_concentration}
\end{equation}
Because $\operatorname{tr}(H(\rho))=1$, the effective dimension of the joint profile is $d(\rho)=\frac{1}{F(\rho)}$.
The first two terms capture concentration within each modality weighted by the squared budget fractions.
The overlap between modalities $\kappa=\operatorname{tr}(H_tH_v)$ is the Frobenius inner product between the trace-normalized text and visual gradient Gram matrices quantified over the same probe samples. 
Positive semidefiniteness implies $\kappa\geq0$, and the Cauchy--Schwarz inequality gives $\kappa\leq1/\sqrt{d_td_v}$.
For fixed $d_t,d_v,\rho$ with $0<\rho<1$, a smaller $\kappa$ reduces $F(\rho)$, whereas $H_t=H_v$ makes the joint profile independent of $\rho$.

\subsection{Prompt allocation from probe gradients}
\label{sec:selection}
\noindentparagraph{\normalfont\bfseries Aggregation of paired probe statistics.}
We repeat the common gradient probe with a fixed configuration, using the same image subsets and pseudo-labels for both modalities. 
For each repetition $r$, we separately normalize the text and visual gradient Gram matrices by their traces to obtain the paired profile $(H_t^{(r)}, H_v^{(r)})$.
We discard the repetitions for which either $\operatorname{tr}(K_t^{(r)})=0$ or $\operatorname{tr}(K_v^{(r)})=0$, and let $R>0$ denote the number of retained repetitions.
The mean concentration within modality $m$ is $c_m=\frac1R\sum_{r=1}^{R}1/{d_m^{(r)}}, m\in\{t,v\}$.
The mean overlap between modalities is $\bar\kappa=\frac1R\sum_{r=1}^{R}
\operatorname{tr}\!\left(H_t^{(r)}H_v^{(r)}\right)$.
The mean spectral concentration of the joint gradient profile across the retained repetitions is 
\begin{equation}
\bar F(\rho)=\frac1R\sum_{r=1}^{R}F^{(r)}(\rho) =c_t\rho^2+c_v(1-\rho)^2
+2\bar\kappa\rho(1-\rho).
\label{eq:aggregate_objective}
\end{equation}
Let $\Omega=c_t+c_v-2\bar\kappa$. $\bar F(\rho)$ is strictly convex when $\Omega>0$, and becomes constant only when $H_t^{(r)}=H_v^{(r)}$ for every retained repetition.

\noindentparagraph{\normalfont\bfseries Continuous budget fraction.}
We minimize $\bar F$ over $[0,1]$ to obtain the continuous text parameter fraction:
\begin{equation}
\rho^*=
\begin{cases}
\displaystyle\operatorname{clip}_{[0,1]}
\!\left(\frac{c_v-\bar\kappa}{\Omega}\right),
&\Omega>0,\\[10pt]
\displaystyle\frac12,&\Omega=0.
\end{cases}
\label{eq:response_optimum}
\end{equation}
Here, $\operatorname{clip}_{[0,1]}$ restricts a scalar to the feasible fraction interval. 
When $\Omega=0$, $H_t^{(r)}=H_v^{(r)}$ for every retained pair, and $\bar F$ is constant over $[0,1]$. 
Therefore, we set $\rho^*=1/2$ at the continuous stage. 
Degenerate cases with zero trace are specified in Appendix~\ref{app:shared_update_analysis}.

The derivative $\bar F'(1/2)=c_t-c_v$ shows that the continuous optimum assigns a larger parameter share to the modality with lower mean concentration. 
The concentration difference and overlap between modalities jointly determine how far the optimum deviates from an equal parameter split. 
When $\bar\kappa=0$, $\rho^*=d_t^{\mathrm{agg}}/(d_t^{\mathrm{agg}}+d_v^{\mathrm{agg}})$, where $d_m^{\mathrm{agg}}=c_m^{-1}$ is the harmonic mean of modality-$m$ effective dimensions across the retained probe repetitions. 

\noindentparagraph{\normalfont\bfseries Projection onto feasible prompt counts.}
The continuous solution assigns parameter budgets $B\rho^*$ and $B(1-\rho^*)$ to the text and visual prompts, respectively. Prompt counts must be integers, and each text token has $h_t$ parameters, whereas each visual token has $h_v$ parameters. 
We project $\rho^*$ onto the feasible allocation set:
\begin{equation}
(\hat n_t,\hat n_v)\in\underset{a\in\mathcal A_B}{\arg\min}
\left|\rho(a)-\rho^*\right|.
\label{eq:target_allocation}
\end{equation}
The resulting text parameter fraction is $\hat\rho=h_t\hat n_t/B$. 
When $\Omega>0$, this projection also minimizes $\bar F(\rho(a))$ over $\mathcal A_B$, as proved in Appendix~\ref{app:shared_update_analysis}. 
When $\Omega=0$, we select the feasible fraction closest to $1/2$. 
If multiple feasible allocations are equally close to $\rho^*$, we choose the one with the smaller text parameter fraction. 
The selected counts satisfy $h_t\hat n_t+h_v\hat n_v=B$, so the selected prompt counts satisfy the exact parameter budget even when $h_t\neq h_v$.

\noindentparagraph{\normalfont\bfseries Optimizing the selected prompts.}
\label{sec:fulltraining}
After allocation, we instantiate text and visual prompts with counts $(\hat n_t,\hat n_v)$ and keep this configuration fixed throughout training. 
For each training image $i$, let $z_i=z(x_i;P_t,P_v)$. 
Using its potentially corrupted observed label $\tilde y_i$, we evaluate the double-softmax loss~\citep{DSPT2026} $\ell(z_i,\tilde y_i)
=-\log\left[\operatorname{softmax} \left(\operatorname{softmax}(z_i)\right)\right]_{\tilde y_i}$.
The selected prompts are optimized by minimizing the mean loss over the full target dataset:
\begin{equation}
\min_{\substack{P_t\in\mathbb R^{\hat n_t\times h_t}\\
P_v\in\mathbb R^{\hat n_v\times h_v}}}
\frac1N\sum_{i=1}^{N}\ell(z_i,\tilde y_i).
\label{eq:robust}
\end{equation}
Only $P_t$ and $P_v$ are updated, while both pretrained encoders remain frozen.
At inference, the predicted class is $\arg\max z_c(x;P_t,P_v)$. 
The gradient probe and prompt allocation are performed only once before training.

\begingroup
\setlength{\parindent}{0pt}
\section{Experiments}
\label{sec:experiments}

\subsection{Experimental setup}
\label{sec:exp_setup}

\begin{table*}[!t]
\centering
\setlength{\tabcolsep}{5.0pt}
\renewcommand{\arraystretch}{0.94}
\caption{Accuracy (\%) comparison with baseline methods under synthetic label noise. GPrompt uses one allocation per dataset under each noise setting. The best results are in \textbf{bold}, and the second-best results are \underline{underlined}.}
\label{tab:sota_comparison}
\label{tab:full_baselines}
\begin{subtable}{\textwidth}
\centering
\caption{Symmetric label noise}
\label{tab:sota_symmetric}
\begin{tabular}{@{}l*{15}{r}@{}}
\toprule
\multicolumn{1}{c}{\multirow{2}{*}{\hspace*{\tabcolsep}Method}} & \multicolumn{3}{c}{Average} & \multicolumn{3}{c}{Caltech101} & \multicolumn{3}{c}{StanfordCars} & \multicolumn{3}{c}{OxfordPets} & \multicolumn{3}{c}{OxfordFlowers} \\
\cmidrule(lr){2-4}\cmidrule(lr){5-7}\cmidrule(lr){8-10}\cmidrule(lr){11-13}\cmidrule(lr){14-16}
& 40\% & 60\% & 80\% & 40\% & 60\% & 80\% & 40\% & 60\% & 80\% & 40\% & 60\% & 80\% & 40\% & 60\% & 80\% \\
\midrule
Zero-shot & 62.42 & 62.42 & 62.42 & 81.43 & 81.43 & 81.43 & 65.33 & 65.33 & 65.33 & 88.19 & 88.19 & 88.19 & 66.19 & 66.19 & 66.19 \\
CoOp & 78.09 & 72.67 & 57.77 & 91.06 & 89.37 & 80.00 & 71.56 & 61.76 & 40.48 & 89.17 & 83.05 & 60.43 & 91.03 & 84.54 & 66.40 \\
Smoothing & 78.39 & 73.22 & 58.81 & 92.82 & 90.54 & 81.29 & 70.58 & 61.45 & 40.90 & 89.38 & 85.85 & 62.76 & \underline{91.89} & 83.89 & 66.91 \\
LogitNorm & 72.32 & 69.61 & 63.48 & 83.06 & 82.14 & 80.21 & 57.05 & 56.97 & 54.05 & 90.75 & 84.18 & 79.02 & 73.11 & 68.96 & 60.82 \\
NLPrompt & 80.56 & 77.43 & 68.82 & \underline{95.36} & 94.42 & 91.85 & \underline{79.70} & \underline{77.21} & \underline{71.98} & 92.41 & 88.97 & 84.18 & \textbf{93.29} & 84.80 & 80.08 \\
DSPT & \underline{81.10} & \underline{79.19} & \underline{74.22} & 95.07 & \underline{95.31} & \underline{94.01} & 79.64 & 76.78 & 71.50 & \underline{93.00} & \underline{92.60} & \underline{91.25} & 91.12 & \underline{88.15} & \underline{80.35} \\
GPrompt & \textbf{82.37} & \textbf{80.91} & \textbf{76.93} & \textbf{96.57} & \textbf{95.98} & \textbf{95.09} & \textbf{80.98} & \textbf{78.72} & \textbf{74.02} & \textbf{94.19} & \textbf{93.69} & \textbf{91.95} & 90.43 & \textbf{89.55} & \textbf{86.06} \\
\midrule
\multicolumn{1}{c}{\multirow{2}{*}{\hspace*{\tabcolsep}Method}} & \multicolumn{3}{c}{Food101} & \multicolumn{3}{c}{FGVCAircraft} & \multicolumn{3}{c}{DTD} & \multicolumn{3}{c}{EuroSAT} & \multicolumn{3}{c}{UCF101} \\
\cmidrule(lr){2-4}\cmidrule(lr){5-7}\cmidrule(lr){8-10}\cmidrule(lr){11-13}\cmidrule(lr){14-16}
& 40\% & 60\% & 80\% & 40\% & 60\% & 80\% & 40\% & 60\% & 80\% & 40\% & 60\% & 80\% & 40\% & 60\% & 80\% \\
\midrule
Zero-shot & 85.46 & 85.46 & 85.46 & 24.37 & 24.37 & 24.37 & 42.73 & 42.73 & 42.73 & 42.91 & 42.91 & 42.91 & 65.19 & 65.19 & 65.19 \\
CoOp & 87.76 & 86.63 & 84.45 & 34.67 & 25.84 & 12.27 & 64.55 & 56.73 & 36.99 & 93.09 & 90.75 & 76.03 & 79.88 & 75.33 & 62.89 \\
Smoothing & 87.71 & 86.76 & 84.66 & 33.33 & 26.02 & 11.80 & 67.77 & 57.15 & 39.95 & 92.03 & 91.00 & 77.41 & 79.97 & 76.31 & 63.57 \\
LogitNorm & 85.32 & 85.61 & 84.73 & 26.73 & 23.07 & 17.56 & 66.84 & 61.81 & 49.07 & 92.30 & 91.33 & 81.00 & 75.73 & 72.43 & 64.89 \\
NLPrompt & 88.50 & 88.08 & 85.78 & 34.74 & 33.38 & 14.86 & 65.97 & 59.17 & 41.25 & 92.04 & 88.83 & 72.01 & 83.00 & \underline{82.04} & \underline{77.40} \\
DSPT & \underline{88.86} & \underline{88.62} & \underline{87.73} & \underline{36.46} & \underline{33.53} & \underline{29.12} & \underline{68.84} & \underline{63.85} & \underline{55.50} & \underline{93.39} & \underline{92.64} & \underline{82.06} & \underline{83.56} & 81.19 & 76.48 \\
GPrompt & \textbf{89.23} & \textbf{88.94} & \textbf{88.29} & \textbf{38.12} & \textbf{34.60} & \textbf{29.38} & \textbf{69.60} & \textbf{65.89} & \textbf{57.93} & \textbf{96.96} & \textbf{96.34} & \textbf{89.81} & \textbf{85.23} & \textbf{84.43} & \textbf{79.82} \\
\bottomrule
\end{tabular}
\end{subtable}
\par\vspace{6pt}
\begin{subtable}{\textwidth}
\centering
\caption{Asymmetric label noise}
\label{tab:sota_asymmetric}
\begin{tabular}{@{}l*{15}{r}@{}}
\toprule
\multicolumn{1}{c}{\multirow{2}{*}{\hspace*{\tabcolsep}Method}} & \multicolumn{3}{c}{Average} & \multicolumn{3}{c}{Caltech101} & \multicolumn{3}{c}{StanfordCars} & \multicolumn{3}{c}{OxfordPets} & \multicolumn{3}{c}{OxfordFlowers} \\
\cmidrule(lr){2-4}\cmidrule(lr){5-7}\cmidrule(lr){8-10}\cmidrule(lr){11-13}\cmidrule(lr){14-16}
& 20\% & 30\% & 40\% & 20\% & 30\% & 40\% & 20\% & 30\% & 40\% & 20\% & 30\% & 40\% & 20\% & 30\% & 40\% \\
\midrule
Zero-shot & 62.42 & 62.42 & 62.42 & 81.43 & 81.43 & 81.43 & 65.33 & 65.33 & 65.33 & 88.19 & 88.19 & 88.19 & 66.19 & 66.19 & 66.19 \\
CoOp & 74.46 & 65.97 & 54.80 & 86.14 & 76.38 & 63.21 & 69.38 & 59.99 & 49.47 & 83.22 & 72.96 & 60.76 & 86.27 & 74.99 & 59.39 \\
Smoothing & 78.02 & 71.12 & 58.89 & 90.39 & 83.77 & 70.47 & 71.61 & 65.21 & 52.51 & 87.97 & 79.41 & 65.58 & \underline{91.09} & 81.44 & 65.28 \\
LogitNorm & 73.61 & 71.08 & 65.31 & 86.44 & 85.27 & 82.71 & 57.96 & 56.49 & 51.48 & 89.73 & 83.46 & 72.27 & 76.01 & 72.89 & 65.15 \\
NLPrompt & 81.56 & \underline{80.79} & 77.73 & 95.56 & 95.21 & 93.63 & 80.90 & 78.48 & 75.01 & \underline{93.63} & \underline{92.85} & 91.27 & 90.37 & \textbf{91.47} & \underline{88.39} \\
DSPT & \underline{81.97} & 80.67 & \underline{79.25} & \underline{96.06} & \underline{95.85} & \underline{93.71} & \underline{81.14} & \underline{78.82} & \underline{75.15} & 93.19 & 92.83 & \underline{91.74} & 91.01 & 86.93 & \textbf{90.07} \\
GPrompt & \textbf{82.65} & \textbf{82.11} & \textbf{80.92} & \textbf{96.67} & \textbf{96.63} & \textbf{95.96} & \textbf{82.60} & \textbf{79.81} & \textbf{77.07} & \textbf{94.40} & \textbf{94.36} & \textbf{93.86} & \textbf{91.29} & \underline{88.40} & 86.73 \\
\midrule
\multicolumn{1}{c}{\multirow{2}{*}{\hspace*{\tabcolsep}Method}} & \multicolumn{3}{c}{Food101} & \multicolumn{3}{c}{FGVCAircraft} & \multicolumn{3}{c}{DTD} & \multicolumn{3}{c}{EuroSAT} & \multicolumn{3}{c}{UCF101} \\
\cmidrule(lr){2-4}\cmidrule(lr){5-7}\cmidrule(lr){8-10}\cmidrule(lr){11-13}\cmidrule(lr){14-16}
& 20\% & 30\% & 40\% & 20\% & 30\% & 40\% & 20\% & 30\% & 40\% & 20\% & 30\% & 40\% & 20\% & 30\% & 40\% \\
\midrule
Zero-shot & 85.46 & 85.46 & 85.46 & 24.37 & 24.37 & 24.37 & 42.73 & 42.73 & 42.73 & 42.91 & 42.91 & 42.91 & 65.19 & 65.19 & 65.19 \\
CoOp & 79.58 & 70.71 & 58.60 & 37.83 & 34.03 & 29.72 & 64.73 & 56.69 & 47.53 & 89.75 & 82.43 & 69.77 & 73.20 & 65.53 & 54.78 \\
Smoothing & 84.01 & 77.77 & 64.24 & 36.98 & 33.27 & 29.28 & 68.60 & 59.98 & 49.94 & 92.56 & 86.99 & 73.65 & 78.99 & 72.26 & 59.06 \\
LogitNorm & 84.71 & 83.86 & 79.05 & 29.01 & 26.41 & 25.26 & 69.43 & 66.50 & 59.30 & 92.86 & 90.91 & 83.31 & 76.31 & 73.90 & 69.30 \\
NLPrompt & \underline{89.05} & 88.65 & 88.41 & 36.71 & 34.45 & \underline{33.60} & 70.17 & \textbf{70.07} & 65.70 & \underline{93.93} & 93.19 & 82.00 & \underline{83.75} & 82.72 & \underline{81.55} \\
DSPT & 89.03 & \underline{88.96} & \underline{88.77} & \textbf{38.46} & \underline{35.93} & \underline{33.60} & \underline{71.03} & \underline{69.93} & \underline{66.74} & \textbf{94.42} & \underline{93.81} & \underline{92.38} & 83.37 & \underline{83.01} & 81.13 \\
GPrompt & \textbf{89.43} & \textbf{89.30} & \textbf{89.18} & \underline{38.03} & \textbf{37.98} & \textbf{36.02} & \textbf{71.92} & \textbf{70.07} & \textbf{69.48} & 93.57 & \textbf{97.02} & \textbf{96.12} & \textbf{85.97} & \textbf{85.45} & \textbf{83.91} \\
\bottomrule
\end{tabular}
\end{subtable}
\end{table*}

\noindent
\textbf{Datasets and label noise.}
Our empirical evaluation is conducted on eleven datasets: nine image classification benchmarks with controlled label noise and two real-world datasets with naturally noisy labels. Specifically, the nine image classification benchmarks comprise Caltech101~\citep{Caltech2004}, StanfordCars~\citep{Cars2013}, OxfordPets~\citep{Pets2012}, OxfordFlowers~\citep{Flowers2008}, Food101~\citep{Food2014}, FGVCAircraft~\citep{Aircraft2013}, DTD~\citep{DTD2014}, EuroSAT~\citep{EuroSAT2019}, and UCF101~\citep{UCF2012}. The two real-world datasets are Clothing1M~\citep{Clothing2015} and WebVision-50~\citep{WebVision2017}. 
For the nine image-classification benchmarks, we split each dataset into training, validation, and test sets following~\citet{CoOp2022}. More details are shown in Table \ref{tab:datasets} in the Appendix. For the label noise settings, given a noise rate $\eta$, we randomly select $\lfloor\eta N\rfloor$ of the $N$ training images for label corruption. We index the labels of the $c$-class target dataset as $\{y_0, y_1, \ldots, y_{c-1}\}$. Under symmetric noise, each selected label $y_i$ is replaced uniformly with one of the other labels in $\{y_j \mid j \in \{0,1,\ldots,c-1\},\; j \neq i\}$. Under asymmetric noise, each selected label $y_i$ is mapped to $y_{(i+1)\bmod c}$. Test labels remain unchanged. Symmetric noise rates are 40\%, 60\%, and 80\%, and asymmetric noise rates are 20\%, 30\%, and 40\%. 
For the experiments with natural label noise, Clothing1M and WebVision-50 retain their original noisy training labels.

\noindent
\textbf{Baselines.}
We compare GPrompt with six baselines.
Zero-shot CLIP uses fixed text prompts without adaptation to the target task~\citep{CLIP2021}. 
CoOp optimizes continuous text prompts with cross entropy loss~\citep{CoOp2022}. 
Smoothing softens the target distribution through label smoothing~\citep{LabelSmoothing2016,LabelSmoothingNoise2020}. LogitNorm regularizes training by normalizing the logit vector~\citep{LogitNorm2022}. 
NLPrompt combines MAE-based prompt learning with sample purification using optimal transport~\citep{NLPrompt2025}. DSPT uses a robust loss to suppress the influence of examples with high prediction errors~\citep{DSPT2026}.

\noindent
\textbf{Training configurations and evaluation measures.}
All experiments use a pretrained CLIP ViT-B/16 backbone~\citep{CLIP2021,ViT2021}. We train the model for 50 epochs using SGD with a learning rate of 0.002, momentum of 0.9, weight decay of $5\times10^{-4}$, and a batch size of 32. We use one warm-up epoch followed by cosine learning-rate decay.
For the nine image-classification benchmarks, we use recognition accuracy (\%) as the evaluation metric. For Clothing1M and WebVision-50, we report Top-1 accuracy (\%) on the clean test and validation sets, respectively.
For each dataset, noise rate, and random seed, all prompt allocations share the same data split, corrupted labels, optimization settings, and parameter budget.
All experiments are repeated three times with different random seeds, and we report the average results.

\noindent
\textbf{Probe and allocation settings.}
\label{sec:implementation}
The candidate prompt counts $(n_t,n_v)$ are $(50,0)$, $(35,10)$, $(20,20)$, and $(5,30)$, each of which exactly satisfies the parameter budget. 
The initialization probe uses a common configuration of 20 text and 20 visual prompt tokens. 
For each initialization seed, we independently sample two subsets of $M=256$ training images, and evaluate each image in its original and horizontally flipped views. 
For each sampled subset, we construct the normalized text and visual gradient matrices, and compute the concentration within each modality and overlap between modalities. 
We average these coefficients over the paired probe repetitions, and derive the budget fraction allocated to text prompts using Eqs.~\ref{eq:response_optimum} and~\ref{eq:target_allocation}. 
For gain ranking, we average $\log(d_t/d_v)$ over the same probe repetitions.
The text and visual responses in each probe are computed from the same initialized model and image subset. We average the concentration and overlap coefficients before computing the allocation, which is then fixed across noise rates and training runs.

\subsection{Overall performance}
\label{sec:baseline_results}

\textbf{Main results.}
We evaluate GPrompt on nine image classification datasets. 
As shown in Table~\ref{tab:sota_comparison}, GPrompt achieves the highest average accuracy across the nine datasets under all six noise settings, outperforming the strongest baselines by 0.68\%--2.71\%.
The improvement is particularly pronounced on EuroSAT, reaching 7.75\% at 80\% symmetric noise.
These results demonstrate that parameter allocation between text and visual prompts for each task consistently strengthens robust prompt learning across diverse downstream tasks and noise conditions.

\textbf{Gains from allocation between text and visual prompts.}
Table~\ref{tab:selected_gain} shows the accuracy gain achieved under GPrompt's selected parameter allocation relative to the text-only configuration (T:V=10:0).
Under symmetric noise, GPrompt improves mean accuracy on all nine datasets, yielding an average accuracy gain of 1.93\% across datasets.
EuroSAT exhibits the largest accuracy gain of 5.60\%. 
Under asymmetric noise, GPrompt achieves improvements on all nine datasets, with an average accuracy gain of 1.87\%.
The improvements are particularly pronounced on EuroSAT (4.06\%), FGVC Aircraft (3.39\%), and Stanford Cars (3.23\%), but remain modest on Caltech101 (0.15\%) and Food101 (0.25\%).
These results demonstrate that the benefit of reallocating prompt parameters varies substantially across tasks, even under an identical training objective and parameter budget.

Table~\ref{tab:natural-noise} reports the mean Top-1 accuracy on the two datasets with natural label noise, Clothing1M and WebVision-50.
GPrompt selects T:V = 4:6 for both datasets, achieving mean accuracies of 68.18\% on Clothing1M and 82.82\% on WebVision-50.
These results exceed the text-only baseline by 1.28\% and 0.94\%, respectively, and are the highest among the four fixed allocations on both datasets.

\begin{table}[t]
\centering
\caption{Accuracy (\%) of text-only prompts (T:V = 10:0) and GPrompt's selected allocation under the same parameter budget and training objective, averaged over three noise rates and three seeds.}
\label{tab:selected_gain}
\setlength{\tabcolsep}{1pt}
\renewcommand{\arraystretch}{1.1}
\newcommand{\selectedgainbase}[1]{#1}
\newcommand{\selectedgainprompt}[2]{$\mathord{#1}_{\scriptscriptstyle(#2)}$}
\begin{tabular*}{\columnwidth}{@{\extracolsep{\fill}}lcccc@{}}
\specialrule{\heavyrulewidth}{0pt}{0.35em}
\multicolumn{1}{c}{\multirow{2}{*}{Dataset}} & \multicolumn{2}{c}{Symmetric} & \multicolumn{2}{c}{Asymmetric} \\
\cmidrule(lr){2-3}\cmidrule(lr){4-5}
 & Text-only & GPrompt & Text-only & GPrompt \\
\specialrule{\lightrulewidth}{0.35em}{0.35em}
Caltech101 & \selectedgainbase{95.75} & \selectedgainprompt{95.88}{+0.13\uparrow} & \selectedgainbase{96.27} & \selectedgainprompt{96.42}{+0.15\uparrow} \\
StanfordCars & \selectedgainbase{74.74} & \selectedgainprompt{77.91}{+3.17\uparrow} & \selectedgainbase{76.59} & \selectedgainprompt{79.82}{+3.23\uparrow} \\
OxfordPets & \selectedgainbase{93.02} & \selectedgainprompt{93.28}{+0.25\uparrow} & \selectedgainbase{93.82} & \selectedgainprompt{94.21}{+0.38\uparrow} \\
OxfordFlowers & \selectedgainbase{85.37} & \selectedgainprompt{88.68}{+3.31\uparrow} & \selectedgainbase{87.36} & \selectedgainprompt{88.81}{+1.44\uparrow} \\
Food101 & \selectedgainbase{88.63} & \selectedgainprompt{88.82}{+0.19\uparrow} & \selectedgainbase{89.05} & \selectedgainprompt{89.30}{+0.25\uparrow} \\
FGVCAircraft & \selectedgainbase{31.19} & \selectedgainprompt{34.03}{+2.84\uparrow} & \selectedgainbase{33.95} & \selectedgainprompt{37.35}{+3.39\uparrow} \\
DTD & \selectedgainbase{64.31} & \selectedgainprompt{64.47}{+0.16\uparrow} & \selectedgainbase{69.39} & \selectedgainprompt{70.49}{+1.10\uparrow} \\
EuroSAT & \selectedgainbase{88.77} & \selectedgainprompt{94.37}{+5.60\uparrow} & \selectedgainbase{91.51} & \selectedgainprompt{95.57}{+4.06\uparrow} \\
UCF101 & \selectedgainbase{81.48} & \selectedgainprompt{83.16}{+1.68\uparrow} & \selectedgainbase{82.32} & \selectedgainprompt{85.11}{+2.79\uparrow} \\
\specialrule{\heavyrulewidth}{0.35em}{0pt}
\end{tabular*}
\end{table}

\subsection{Ablation study}

Table~\ref{tab:normalization-ablation} compares GPrompt with three variants: text-only prompting, random allocation, and allocation without trace normalization. 
For random allocation, we report the expected accuracy of random selection among the four fixed parameter ratios. 
The ablation results yield three observations. 
First, removing multimodal prompting (w/o \textit{Mm. Prom.}) causes the largest accuracy drop across all three datasets, highlighting the benefit of jointly learning text and visual prompts under label noise.
Second, replacing GPrompt's selected allocation with random allocation (w/o \textit{Grad. Probe}) decreases mean accuracy by 1.32\%, validating the importance of selecting suitable parameter allocation for each task. Third, removing trace normalization (w/o \textit{Trace Norm.}) reduces accuracy by 1.40\%, showing that normalization prevents gradient-scale differences across modalities from biasing allocation decision.

\subsection{Analysis of allocation selection}
\label{sec:allocation_results}

GPrompt selects parameter allocations for each task before training any candidate configurations. 
Taking OxfordFlowers and UCF101 as examples, Figure~\ref{fig:allocation_curves} illustrates accuracy across different parameter allocations between text and visual prompts under symmetric and asymmetric label noise. The optimal allocation varies across datasets. Under asymmetric noise, the optimal parameter ratios between text and visual prompts are 4:6 for OxfordFlowers and 7:3 for UCF101. These results highlight the limitations of existing methods that rely on fixed parameter allocations between text and visual prompts. 
More importantly, Figure~\ref{fig:allocation_curves} shows that GPrompt generally selects parameter allocations suited to each task and achieves the highest accuracy on both datasets under nearly all noise settings.

\begin{table}[t]
\centering
\setlength{\tabcolsep}{3pt}
\renewcommand{\arraystretch}{1.1}
\caption{Top-1 accuracy (\%) with different parameter allocations on two datasets with natural label noise. \colorbox{gray!25}{Gray} shading indicates the Top-1 accuracy achieved using the parameter ratio selected by GPrompt. The best results are in \textbf{bold}, and the second-best results are \underline{underlined}.}

\label{tab:natural-noise}
\begin{tabular*}{\columnwidth}{@{\extracolsep{\fill}}ccccc@{}}
\toprule
\multirow[c]{2}{*}{Dataset} & \multicolumn{4}{c}{Parameter ratio (T:V)} \\
\cmidrule(l){2-5}
& 10:0 & 7:3 & 4:6 & 1:9 \\
\midrule
Clothing1M & $66.90$ & $67.74_{(+0.84\uparrow)}$ & \cellcolor{gray!25}$\mathbf{68.18}_{(+1.28\uparrow)}$ & $\underline{67.81}_{(+0.91\uparrow)}$ \\
WebVision-50 & $81.87$ & $\underline{82.48}_{(+0.61\uparrow)}$ & \cellcolor{gray!25}$\mathbf{82.82}_{(+0.95\uparrow)}$ & $81.86_{(-0.01\downarrow)}$ \\
\bottomrule
\end{tabular*}
\end{table}

\begin{figure}[!t]
\centering
\includegraphics[width= 0.95\linewidth]{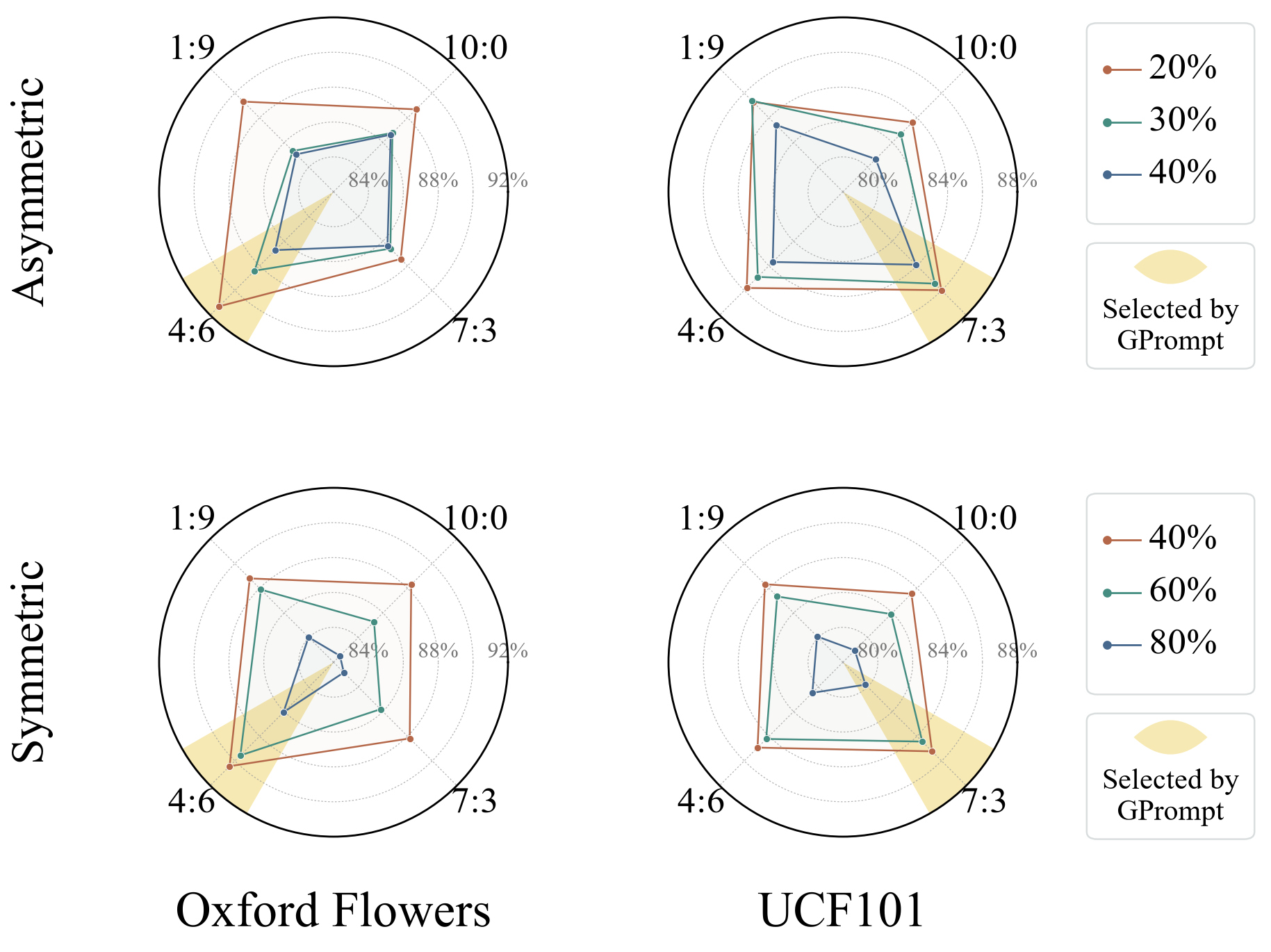}
\caption{Accuracy (\%) across four fixed parameter allocations between text and visual prompts and the ratio selected by GPrompt on two datasets under asymmetric (top) and symmetric (bottom) label noise.}
\Description{Four radar panels show Oxford Flowers on the left and UCF101 on the right, with asymmetric noise above and symmetric noise below. Yellow sectors highlight the allocations selected by GPrompt: T:V = 4:6 for Oxford Flowers and 7:3 for UCF101 in both rows.}
\label{fig:allocation_curves}
\label{fig:allocation_asym}
\label{fig:allocation_sym}
\end{figure}

\begin{figure}[!t]
\centering
\includegraphics[width=0.95\linewidth]{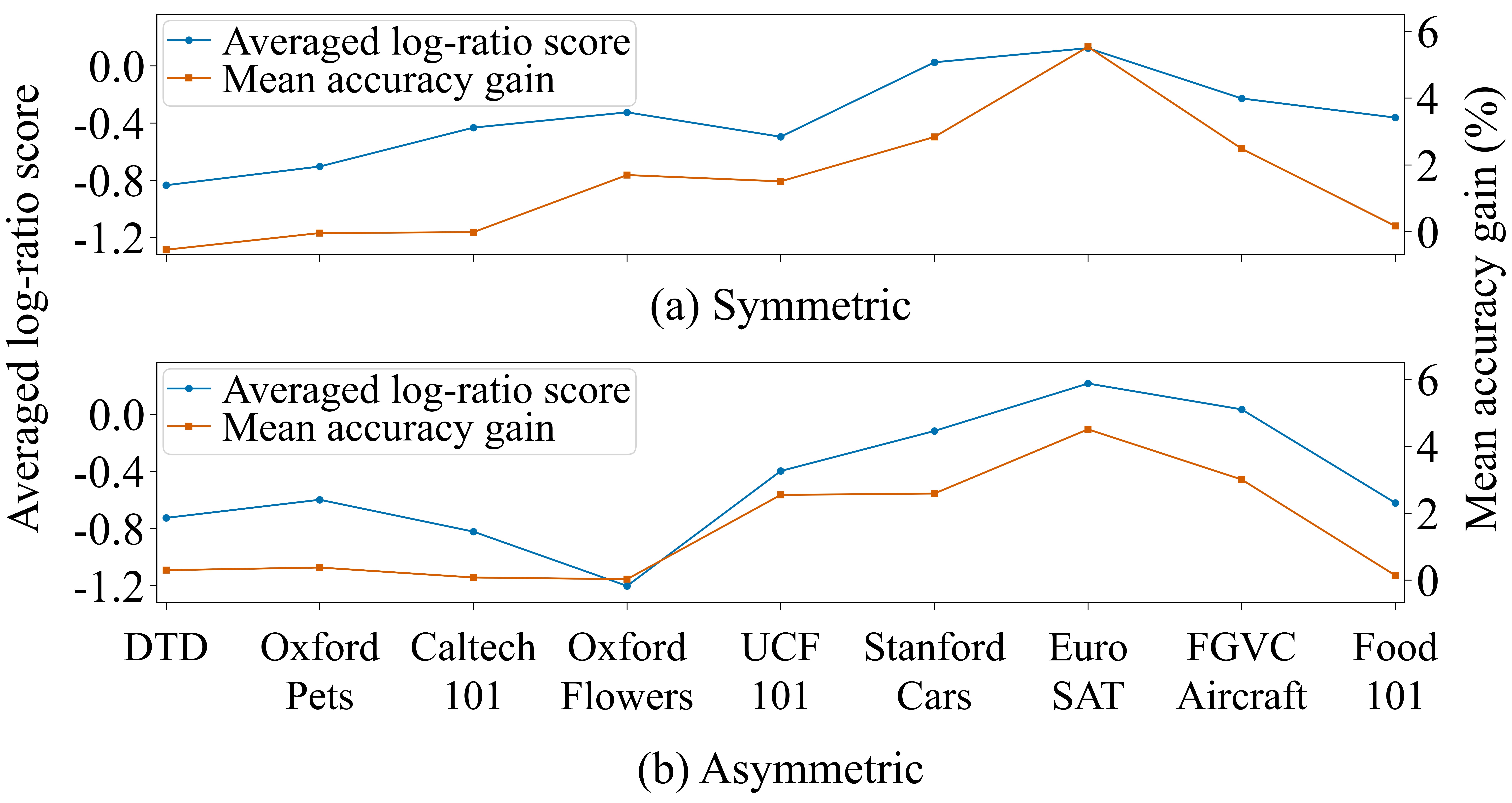}
\caption{The averaged log-ratio score $\bar s_{\mathrm{gain}}$ and mean accuracy gain $b_u$ from visual prompts on nine datasets under symmetric (top) and asymmetric (bottom) label noise.}
\Description{Two vertically stacked dual-axis line plots compare the mean log-ratio probe score with mean accuracy gains from visual prompts across nine datasets. Symmetric noise is shown above and asymmetric noise below; the two vertical axes show probe score and accuracy gain.}
\label{fig:gain_prediction}
\end{figure}

Table~\ref{tab:oracle_summary} compares three oracle policies over fully trained candidate allocations.
Within each dataset and noise type (symmetric or asymmetric), $Oracle_{dataset}$ selects the allocation with the highest accuracy averaged over noise rates and random seeds. $Oracle_{noise~rate}$ selects the allocation with the highest mean accuracy across the three random seeds separately at each noise rate. $Oracle_{run}$ selects the best allocation independently for each dataset, noise rate, and random seed within each noise type.
Under the symmetric noise setting, GPrompt achieves an accuracy of 80.07\%, falling short of $Oracle_{dataset}$ and $Oracle_{noise~rate}$ by only 0.02\% and 0.09\%, respectively. Under the asymmetric noise setting, GPrompt achieves an accuracy of 81.90\%, with corresponding gaps of only 0.13\% and 0.20\%, respectively. Moreover, GPrompt correctly matches the oracle parameter allocation for 5 of the 9 datasets under symmetric noise and 7 of the 9 datasets under asymmetric noise.

\begin{table}[t]
\centering
\caption{Experimental results of ablation study. We report Accuracy (\%) on Oxford Flowers, Stanford Cars, and EuroSAT, averaged across the evaluated symmetric and asymmetric label-noise settings. The best results are in \textbf{bold}, and the second-best results are \underline{underlined}.}
\label{tab:normalization-ablation}
\begingroup
\setlength{\tabcolsep}{3.0pt}
\renewcommand{\arraystretch}{1.1}
\setlength{\dashlinedash}{4pt}\setlength{\dashlinegap}{2pt}
\begin{tabular*}{\columnwidth}{@{\extracolsep{\fill}}lccc@{}}
\toprule
\multicolumn{1}{c}{\makecell[c]{Method}}
  & \makecell[c]{Oxford\\Flowers}
  & \makecell[c]{Stanford\\Cars}
  & EuroSAT \\
\midrule
\textbf{GPrompt (ours)}
  & \textbf{88.75}
  & \textbf{78.87}
  & \textbf{94.97} \\
\addlinespace[0.5pt]
\hdashline
\addlinespace[1pt]
- w/o \textit{Mm. Prom.}
  & $86.36_{(-2.39\downarrow)}$
  & $75.66_{(-3.21\downarrow)}$
  & $90.14_{(-4.83\downarrow)}$ \\
- w/o \textit{Grad. Probe}
  & $\underline{87.01}_{(-1.74\downarrow)}$
  & $\underline{77.70}_{(-1.17\downarrow)}$
  & $93.91_{(-1.06\downarrow)}$ \\
- w/o \textit{Trace Norm.}
  & $86.73_{(-2.02\downarrow)}$
  & $77.48_{(-1.39\downarrow)}$
  & $\underline{94.19}_{(-0.78\downarrow)}$ \\
\bottomrule
\end{tabular*}
\endgroup
\end{table}

\begin{table}[t]
\centering
\caption{Experimental results for the four fixed parameter ratios and GPrompt, compared with $Oracle_{dataset}$. Accuracy (\%) denotes the accuracy averaged across the nine image classification benchmarks under different noise rates. Hit indicates the number of datasets for which a method's allocation matches that of $Oracle_{dataset}$. The best results are shown in \textbf{bold}, and the second-best results are \underline{underlined}.}
\label{tab:policies}
\label{tab:oracle_summary}

\setlength{\tabcolsep}{3.0pt}
\renewcommand{\arraystretch}{1.1}
\setlength{\dashlinedash}{4pt}\setlength{\dashlinegap}{2pt}
\begin{tabular*}{\columnwidth}{@{\extracolsep{\fill}}lcccc@{}}
\toprule
\multicolumn{1}{c}{\multirow{2}{*}{Policy}} & \multicolumn{2}{c}{Symmetric} & \multicolumn{2}{c}{Asymmetric} \\
\cmidrule(lr){2-3}\cmidrule(lr){4-5}
 & Accuracy (\%) & Hit & Accuracy (\%) & Hit \\
\midrule
$Oracle_{run}$ & 80.35 & -- & 82.27 & -- \\
$Oracle_{noise~rate}$ & 80.16 & -- & 82.10 & -- \\
$Oracle_{dataset}$ & 80.09 & -- & 82.03 & -- \\
\addlinespace[0.5pt]
\hdashline
\addlinespace[1pt]
Fixed, T:V = 10:0 & 78.14{\scriptsize$(-1.95\mathord{\downarrow})$} & 0/9 & 80.03{\scriptsize$(-2.00\mathord{\downarrow})$} & 0/9 \\
Fixed, T:V = 7:3 & 79.63{\scriptsize$(-0.46\mathord{\downarrow})$} & 2/9 & 81.73{\scriptsize$(-0.30\mathord{\downarrow})$} & 5/9 \\
Fixed, T:V = 4:6 & \underline{79.92}{\scriptsize$(-0.16\mathord{\downarrow})$} & 5/9 & \underline{81.75}{\scriptsize$(-0.28\mathord{\downarrow})$} & 2/9 \\
Fixed, T:V = 1:9 & 79.42{\scriptsize$(-0.67\mathord{\downarrow})$} & 2/9 & 81.14{\scriptsize$(-0.89\mathord{\downarrow})$} & 2/9 \\
\textbf{GPrompt (ours)} & \textbf{80.07}{\scriptsize$(-0.02\mathord{\downarrow})$} & 5/9 & \textbf{81.90}{\scriptsize$(-0.13\mathord{\downarrow})$} & 7/9 \\
\bottomrule
\end{tabular*}
\end{table}

\subsection{Predicting gains from visual prompts}
\label{sec:probe_results}

For each dataset and noise setting, we define $b_u$ as the mean accuracy gain of three allocations containing visual prompts (T:V = 7:3, 4:6, and 1:9) relative to the text-only configuration (T:V = 10:0), averaged across three noise rates. 
Figure~\ref{fig:gain_prediction} compares the averaged log-ratio score $\bar s_{\mathrm{gain}}$ with $b_u$. Spearman correlations are 0.950 under symmetric noise and 0.983 under asymmetric noise, indicating that higher probe log-ratio scores are consistently associated with larger accuracy gains from visual prompts under both noise settings.

\subsection{Cost of allocation assessment}
\label{sec:efficiency_results}

To assess the computational cost of GPrompt's gradient probe for determining the parameter allocation between text and visual prompts, we compare runtime of GPrompt's gradient probe with that of exhaustive allocation search, which entails fully training and evaluating all candidate allocations. As shown in Figure~\ref{fig:runtime}, across the nine benchmarks, GPrompt's gradient probe takes only 38–69 seconds, substantially less than the 3,342–50,100 seconds required for the exhaustive allocation search. In particular, on Food101, GPrompt's gradient probe saves approximately 50,058 seconds. 
Moreover, across all eleven datasets, the gradient probe reduces the allocation assessment time by 99.9\% relative to exhaustive allocation search. Table~\ref{tab:selection_cost} further demonstrates the end-to-end training-time savings enabled by the gradient probe, yielding a 3.48$\times$–3.94$\times$ speedup over exhaustive allocation search. These results demonstrate GPrompt's substantial computational efficiency for allocation selection.

\begin{table}[t]
\centering
\caption{End-to-end training time (min) for GPrompt and exhaustive allocation search. Speedup is computed as the ratio of the end-to-end runtime of exhaustive allocation search to that of GPrompt.}
\label{tab:timing}
\label{tab:selection_cost}

\begingroup
\setlength{\tabcolsep}{3pt}
\renewcommand{\arraystretch}{1.05}
\begin{tabular*}{\columnwidth}{@{\extracolsep{\fill}}lccc@{\hspace{6pt}}}
\toprule
\multicolumn{1}{@{}c}{Dataset} & GPrompt & Allo. Search & speedup \\
\midrule
Caltech101 & 18.5 & 66.3 & 3.58$\times$ \\
DTD & 14.8 & 55.7 & 3.76$\times$ \\
EuroSAT & 48.7 & 169.3 & 3.48$\times$ \\
FGVCAircraft & 53.4 & 210.6 & 3.94$\times$ \\
Food101 & 217.5 & 835.0 & 3.84$\times$ \\
OxfordFlowers & 21.3 & 81.9 & 3.85$\times$ \\
OxfordPets & 17.7 & 67.4 & 3.81$\times$ \\
StanfordCars & 56.8 & 217.9 & 3.84$\times$ \\
UCF101 & 26.9 & 105.6 & 3.93$\times$ \\
\bottomrule
\end{tabular*}
\endgroup
\end{table}
\begin{figure}[!t]
\centering
\includegraphics[width=0.91\linewidth]{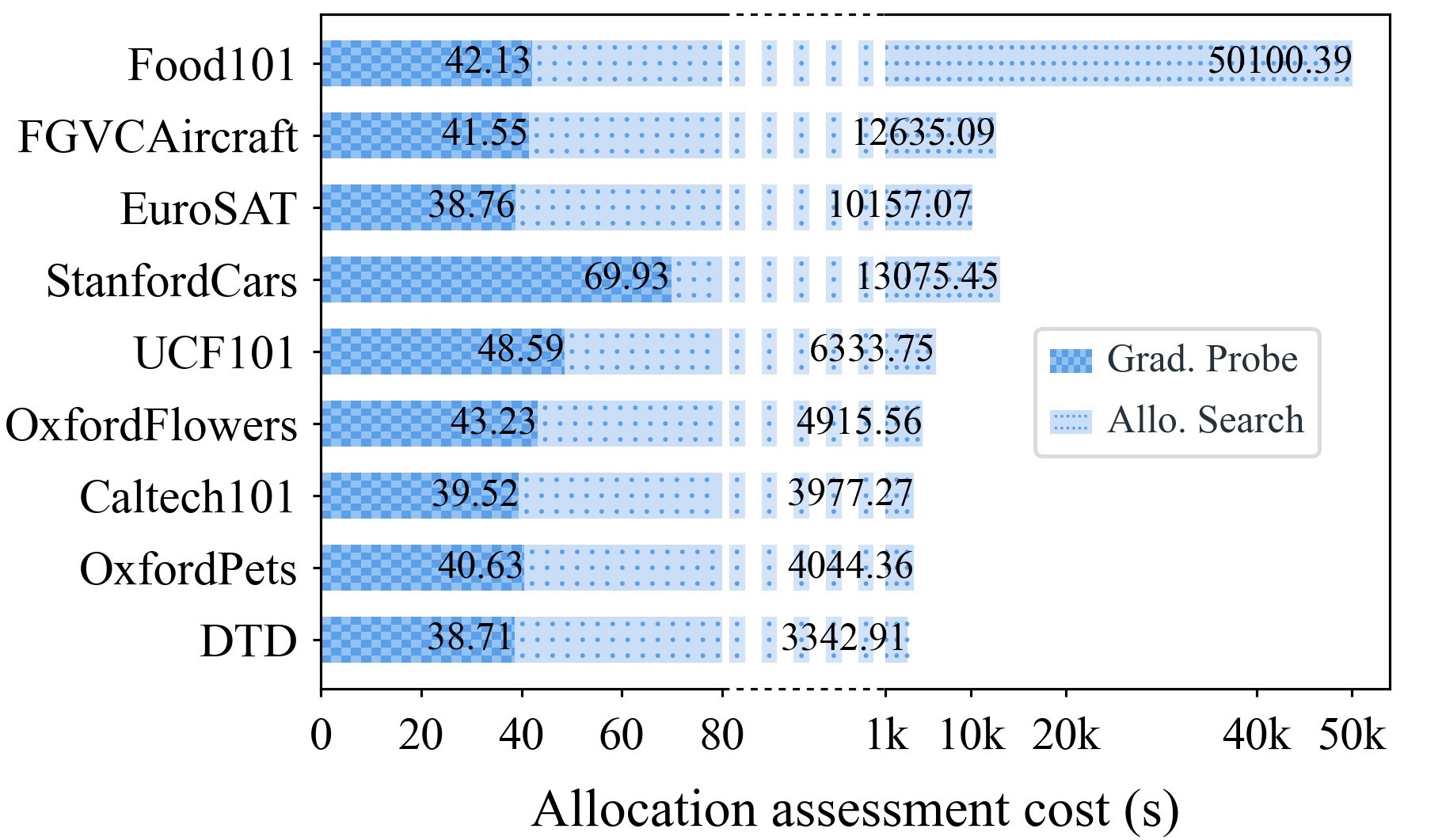}
\caption{Allocation-required time (s) comparison between GPrompt’s gradient probe (Grad. Probe) and the exhaustive allocation
search (Allo. Search) across nine image classification benchmarks with asymmetric label noise.}
\Description{Horizontal bars compare gradient-probe time with exhaustive four-candidate allocation-search time in seconds across nine image classification datasets under asymmetric noise. A broken horizontal axis displays the different time scales.}
\label{fig:runtime}
\end{figure}

\par
\endgroup

\section{Conclusion}
\label{sec:conclusion}

We introduced GPrompt, a training-free framework that allocates trainable parameters between text and visual prompts for robust prompt learning under label noise. 
GPrompt probes paired gradients at initialization to characterize intra-modality concentration and inter-modality overlap. 
It then minimizes the spectral concentration of joint gradient profile to derive a closed-form budget allocation, which is projected onto feasible prompt counts satisfying the exact parameter constraint. 
GPrompt trains only the selected configuration with both pretrained encoders frozen, avoiding repeated candidate training while enhancing robustness. 
Across eleven datasets spanning symmetric, asymmetric, and natural label noise, GPrompt outperforms the strongest baseline by up to 7.75\%. Notably, under asymmetric noise, GPrompt achieves mean accuracy within 0.13\% of the best allocation identified via exhaustive training. 
Moreover, the gradient probe reduces allocation-assessment time by 99.9\% relative to exhaustive candidate training. 
Together, these experimental results demonstrate the effectiveness of task-adaptive prompt allocation for improving robustness under noisy supervision.

\label{end:main}
\clearpage
\bibliographystyle{ACM-Reference-Format}
\bibliography{references}
\appendix
\section{Additional experimental results}
\label{app:additional_results}

\subsection{Allocation examples}
\label{app:allocation_curves}
Figure~\ref{fig:allocation_curves} in the main text shows four allocation examples: Oxford Flowers on the left and UCF101 on the right, with asymmetric noise in the top row and symmetric noise in the bottom row. Each panel compares the four candidate allocations. Natural-noise results are presented separately in Table~\ref{tab:natural-noise}.

\subsection{Allocation accuracy without candidate training}
\label{app:policy_summary}
GPrompt selects directly from initialization gradients, matching $Oracle_{dataset}$ on five of nine datasets under symmetric noise and seven of nine under asymmetric noise. Table~\ref{tab:policies} reports mean accuracy and hits against $Oracle_{dataset}$ for GPrompt and fixed policies, alongside three Oracle accuracy references. Fixed policies use one ratio across tasks.

\subsection{Per-dataset allocation and Oracle results}
\label{app:oracle_details}
Table~\ref{tab:per_dataset} reports selected allocations, paired gains, and all three Oracle references for every dataset. Results average the three noise rates within each seed before aggregation across seeds.

\begin{table*}[t]
\centering
\caption{Results for each dataset under symmetric (left) and asymmetric (right) noise. Accuracy and paired gain are averaged over three noise rates within each seed and then summarized by the mean and sample standard deviation across three seeds. Gap is Oracle accuracy of per dataset minus GPrompt's accuracy, in \%. Natural-noise results appear separately in Table~\ref{tab:natural-noise}.}
\label{tab:per_dataset}
\label{tab:symmetric_capacity}
\label{tab:gprompt_asym}

\begingroup
\small
\setlength{\tabcolsep}{2.5pt}
\renewcommand{\arraystretch}{1.05}
\begin{tabular*}{\textwidth}{@{\extracolsep{\fill}}lcccccccc@{\hspace{8pt}}cccccccc@{}}
\toprule
\multicolumn{1}{c}{\multirow[c]{4.7}{*}{Dataset}} & \multicolumn{8}{c}{Symmetric noise} & \multicolumn{8}{c}{Asymmetric noise} \\
\cmidrule(lr){2-9}\cmidrule(lr){10-17}
 & \multicolumn{2}{c}{T:V} & \multicolumn{2}{c}{GPrompt} & \multicolumn{3}{c}{Oracle accuracy} & \multirow[c]{3.3}{*}{Gap}
 & \multicolumn{2}{c}{T:V} & \multicolumn{2}{c}{GPrompt} & \multicolumn{3}{c}{Oracle accuracy} & \multirow[c]{3.3}{*}{Gap} \\
\cmidrule(lr){2-3}\cmidrule(lr){4-5}\cmidrule(lr){6-8}\cmidrule(lr){10-11}\cmidrule(lr){12-13}\cmidrule(lr){14-16}
 & Selected & Oracle & Acc. & Gain & \makecell{per\\dataset} & \makecell{per\\rate} & \makecell{per\\run} &
 & Selected & Oracle & Acc. & Gain & \makecell{per\\dataset} & \makecell{per\\rate} & \makecell{per\\run} & \\
\midrule
Caltech101 & 1:9 & 1:9 & 95.88$_{\pm0.10}$ & 0.13$_{\pm0.30}$ & 95.88 & 95.91 & 96.09 & 0.00 & 1:9 & 1:9 & 96.42$_{\pm0.16}$ & 0.15$_{\pm0.12}$ & 96.42 & 96.50 & 96.60 & 0.00 \\
DTD & 7:3 & 7:3 & 64.47$_{\pm1.17}$ & 0.16$_{\pm0.87}$ & 64.47 & 64.47 & 64.91 & 0.00 & 7:3 & 7:3 & 70.49$_{\pm0.58}$ & 1.10$_{\pm0.56}$ & 70.49 & 70.71 & 70.89 & 0.00 \\
EuroSAT & 4:6 & 1:9 & 94.37$_{\pm0.36}$ & 5.60$_{\pm0.35}$ & 94.38 & 94.50 & 94.65 & 0.01 & 4:6 & 7:3 & 95.57$_{\pm2.00}$ & 4.06$_{\pm1.83}$ & 96.75 & 96.83 & 96.94 & 1.18 \\
FGVC Aircraft & 4:6 & 4:6 & 34.03$_{\pm1.06}$ & 2.84$_{\pm1.27}$ & 34.03 & 34.08 & 34.53 & 0.00 & 4:6 & 4:6 & 37.35$_{\pm1.16}$ & 3.39$_{\pm1.20}$ & 37.35 & 37.45 & 37.71 & 0.00 \\
Food101 & 7:3 & 4:6 & 88.82$_{\pm0.09}$ & 0.19$_{\pm0.12}$ & 88.86 & 88.87 & 88.89 & 0.04 & 7:3 & 7:3 & 89.30$_{\pm0.03}$ & 0.25$_{\pm0.03}$ & 89.30 & 89.30 & 89.33 & 0.00 \\
Oxford Flowers & 4:6 & 4:6 & 88.68$_{\pm0.95}$ & 3.31$_{\pm0.59}$ & 88.68 & 88.68 & 88.72 & 0.00 & 4:6 & 4:6 & 88.81$_{\pm0.19}$ & 1.44$_{\pm0.70}$ & 88.81 & 88.81 & 89.36 & 0.00 \\
Oxford Pets & 7:3 & 7:3 & 93.28$_{\pm0.46}$ & 0.25$_{\pm0.29}$ & 93.28 & 93.30 & 93.46 & 0.00 & 7:3 & 1:9 & 94.21$_{\pm0.10}$ & 0.38$_{\pm0.17}$ & 94.21 & 94.27 & 94.32 & 0.00 \\
Stanford Cars & 7:3 & 4:6 & 77.91$_{\pm0.23}$ & 3.17$_{\pm0.28}$ & 77.98 & 78.22 & 78.26 & 0.07 & 7:3 & 7:3 & 79.82$_{\pm0.06}$ & 3.23$_{\pm0.16}$ & 79.82 & 79.90 & 79.91 & 0.00 \\
UCF101 & 7:3 & 4:6 & 83.16$_{\pm0.47}$ & 1.68$_{\pm0.69}$ & 83.21 & 83.38 & 83.61 & 0.05 & 7:3 & 7:3 & 85.11$_{\pm0.52}$ & 2.79$_{\pm0.53}$ & 85.11 & 85.11 & 85.42 & 0.00 \\
\midrule
Average & -- & -- & 80.07$_{\pm0.21}$ & 1.93$_{\pm0.06}$ & 80.09 & 80.16 & 80.35 & 0.02 & -- & -- & 81.90$_{\pm0.25}$ & 1.87$_{\pm0.32}$ & 82.03 & 82.10 & 82.27 & 0.13 \\
\bottomrule
\end{tabular*}
\endgroup
\end{table*}

\subsection{Per-dataset computation costs}
\label{app:runtime_details}
Table~\ref{tab:timing} gives the wall-clock measurements underlying Figure~\ref{fig:runtime}. The time for two probe calls is twice the mean time for one probe call. Training includes optimization and test evaluation.

For an additional summary across eleven datasets, we include Clothing1M and WebVision-50 and use recorded active training plus evaluation durations. For each dataset, we average six recorded durations of individual probe calls. For the four-candidate reference, we sum candidate costs within each seed, then average over three seeds and, for the nine controlled-noise datasets, three asymmetric-noise rates. Summing these means for the individual datasets gives 436.60\,s for one probe call and 636412.76\,s for training and evaluating all four candidates, a 99.9\% reduction. The probe numerator represents one call per dataset. Hardware and concurrency vary across batches; these values are a descriptive aggregate of recorded durations. The historical nine-dataset wall-clock figure and table retain their original measurement definitions.

\begin{table}[t]
\caption{Datasets used in our experiments, grouped by the source of label noise.}
\label{tab:datasets}
\label{tab:web_datasets}
\centering
\small
\setlength{\tabcolsep}{4pt}
\begin{tabular}
{@{}p{0.46\linewidth}rp{0.31\linewidth}@{}}
\toprule
Dataset & Classes & Domain \\
\midrule
\multicolumn{3}{@{}l}{\textit{Synthetic label noise}} \\
Caltech101~\citep{Caltech2004} & 100 & Objects \\
DTD~\citep{DTD2014} & 47 & Textures \\
EuroSAT~\citep{EuroSAT2019} & 10 & Satellite imagery \\
FGVC Aircraft~\citep{Aircraft2013} & 100 & Aircraft \\
Food101~\citep{Food2014} & 101 & Food \\
Oxford Flowers~\citep{Flowers2008} & 102 & Flowers \\
Oxford Pets~\citep{Pets2012} & 37 & Pets \\
Stanford Cars~\citep{Cars2013} & 196 & Cars \\
UCF101~\citep{UCF2012} & 101 & Human actions \\
\midrule
\multicolumn{3}{@{}l}{\textit{Natural label noise}} \\
Clothing1M~\citep{Clothing2015} & 14 & Clothing \\
WebVision-50~\citep{WebVision2017} & 50 & Web images \\
\bottomrule
\end{tabular}
\par\smallskip
\raggedright\footnotesize Caltech101 excludes the background and Faces\_easy categories; UCF101 uses video midframes. Clothing1M uses 1,000,000 noisy training and 10,526 clean test images. WebVision-50 uses 65,944 Google training images (classes 0--49) and 2,500 clean validation images.
\end{table}

\section{Analysis of the analytic allocation}
\label{app:shared_update_analysis}

\subsection{Response concentration}
For a nonzero operator, $H_m=K_m/T_m$ has nonnegative eigenvalues $\pi_{m,k}$ summing to one. Its squared concentration is $\|H_m\|_F^2=\sum_k\pi_{m,k}^2=1/d_m$. For any orthogonal projection $Q$ of rank $k$,
\begin{equation}
\operatorname{tr}(QH_m)\leq\|Q\|_F\|H_m\|_F
=\sqrt{k/d_m}.
\label{eq:mode_coverage}
\end{equation}
Retaining a fraction $\zeta\in(0,1]$ of the normalized spectral mass requires $k\geq\zeta^2d_m$. For a fixed retained fraction, lower concentration gives a larger lower bound on the required subspace dimension.

\subsection{The gradient profile per unit of budget}
We construct the joint gradient profile for probed parameters. Let $b_{m,p}\in\mathbb R^M$ denote column $p$ of $G_m$. For its nonzero columns, define normalized direction and sampling probability as
$u_{m,p}=\frac{b_{m,p}}{\|b_{m,p}\|},
\nu_{m,p}=\frac{\|b_{m,p}\|^2}{T_m}.$
Because $T_m=\sum_p\|b_{m,p}\|^2$, these probabilities sum to one. Drawing a direction $u_m$ from this empirical distribution gives
\begin{equation}
\mathbb E[u_mu_m^\top]
=\sum_p\nu_{m,p}u_{m,p}u_{m,p}^\top
=\frac{G_mG_m^\top}{T_m}=H_m.
\label{eq:expected_profile}
\end{equation}
Thus, $H_m$ is the mean profile of a unit response direction, with the direction distribution weighted by its observed gradient norms.

For a candidate allocation $a$ with text-budget fraction $\rho$, the normalized joint gradient geometry is
\begin{equation}
\begin{aligned}
\widetilde H_a={}&\frac1B\left[
\sum_{j=1}^{B_t}u_{t,j}u_{t,j}^\top
+\sum_{j=1}^{B_v}u_{v,j}u_{v,j}^\top\right],\\
\mathbb E[\widetilde H_a]={}&
\rho(a)H_t+(1-\rho(a))H_v.
\end{aligned}
\label{eq:budget_profile_derivation}
\end{equation}
The draws here define the surrogate, rather than an additional sampling step during prompt training. The underlying assumption is that the empirical distribution of directions represents a modality's normalized profile across the selected prompt counts. We then optimize concentration of the expected profile.

For fixed probe parameters, the same operator has a first-order interpretation under the following normalized block updates:
\begin{equation}
\begin{aligned}
\delta\theta_t&=-\eta\frac{\rho}{T_t}G_t^\top w,\\
\delta\theta_v&=-\eta\frac{1-\rho}{T_v}G_v^\top w,\\
\Delta\boldsymbol\ell&=-\eta[\rho H_t+(1-\rho)H_v]w
+O(\eta^2).
\end{aligned}
\label{eq:normalized_joint_step}
\end{equation}
The two block responses add at first order, while cross derivatives enter the second order remainder.

\subsection{Overlap and the analytic optimum}
Both modality operators act on the same sampled losses, although their parameter dimensions can differ. Their overlap admits the equivalent expressions
\begin{equation}
\begin{aligned}
\kappa&=\operatorname{tr}(H_tH_v)
=\frac{\|G_t^\top G_v\|_F^2}{T_tT_v},\\
\omega&=\frac{\kappa}{\|H_t\|_F\|H_v\|_F}
=\kappa\sqrt{d_td_v}\in[0,1].
\end{aligned}
\label{eq:response_overlap}
\end{equation}
The Gram form proves nonnegativity, and the Frobenius inner product inequality gives the upper bound. Zero overlap means the response subspaces are orthogonal. $\omega=1$ means the unit trace operators coincide.

Averaging Eq.~\ref{eq:joint_concentration} gives
\begin{equation}
\begin{aligned}
\bar F(\rho)&=c_v+2(\bar\kappa-c_v)\rho+\Omega\rho^2,\\
\bar F'(\rho)&=2(\bar\kappa-c_v)+2\Omega\rho,\\
\Omega&=\frac1R\sum_r\|H_t^{(r)}-H_v^{(r)}\|_F^2.
\end{aligned}
\label{eq:concentration_derivative}
\end{equation}
For $\Omega>0$, solving $\bar F'(\rho)=0$ and restricting the solution to $[0,1]$ proves Eq.~\ref{eq:response_optimum}. An interior optimum occurs exactly when $\bar\kappa<\min(c_t,c_v)$. Its concentration is
\begin{equation}
\bar F(\rho^*)=\frac{c_tc_v-\bar\kappa^2}{c_t+c_v-2\bar\kappa}.
\label{eq:optimal_concentration}
\end{equation}
For distinct profiles, $\bar\kappa\geq c_v$ gives $\rho^*=0$, and $\bar\kappa\geq c_t$ gives $\rho^*=1$. The optimum therefore depends jointly on the two concentrations and their overlap.

The unconstrained solution $\rho_0=(c_v-\bar\kappa)/\Omega$ obeys
\begin{equation}
\rho_0-\tfrac12
=\frac{c_v-c_t}{2\Omega}
=\frac{d_t^{\mathrm{agg}}-d_v^{\mathrm{agg}}}
{2d_t^{\mathrm{agg}}d_v^{\mathrm{agg}}\Omega}.
\label{eq:optimum_contrast}
\end{equation}
This separates the direction of preference from its magnitude. Orthogonal profiles give the harmonic dimension allocation given in Section~\ref{sec:selection}. Their combined effective dimension at the optimum is $d_t^{\mathrm{agg}}+d_v^{\mathrm{agg}}$, when effective dimension is measured as the reciprocal of mean concentration. If all paired profiles coincide, $\Omega=0$ and the objective is constant.

\subsection{Discrete allocation under the exact budget}
For $\Omega>0$, completing the square gives
\begin{equation}
\bar F(\rho)=\Omega(\rho-\rho_0)^2
+c_v-\frac{(c_v-\bar\kappa)^2}{\Omega}.
\label{eq:projection_proof}
\end{equation}
When $\rho_0\in[0,1]$, minimizing this expression on $\{\rho(a):a\in\mathcal A_B\}$ is equivalent to choosing the nearest fraction to $\rho^*=\rho_0$. If $\rho_0<0$, the objective increases on $[0,1]$, so both minimization and projection to $\rho^*=0$ choose the smallest feasible fraction. If $\rho_0>1$, both choose the largest. For $\Omega=0$, every candidate minimizes the objective, and the rule chooses the one closest to one half. The smaller text fraction resolves an equal distance.

Consequently, the discrete allocation in Eq.~\ref{eq:target_allocation} satisfies
\begin{equation}
\hat a\in\arg\min_{a\in\mathcal A_B}\bar F(\rho(a)).
\label{eq:discrete_optimality}
\end{equation}
The fixed budget constraint ensures that the token counts jointly satisfy the parameter budget. Independently rounding the two continuous token counts may violate this constraint.

\subsection{Degenerate probe conventions}
The normalized objective uses paired repetitions in which both text and visual gradient matrices are nonzero. If some repetitions have a zero block, we omit those pairs from the normalized objective and let $R$ denote the retained count. For a retained pair, no denominator floor is added to $H_m$ or $d_m$.

If no pair is retained, let $E_t,E_v$ be the sums of the respective block energies over all attempted repetitions. Use the fallback fraction
\begin{equation}
\rho^*=
\begin{cases}
1,&E_t>0,\ E_v=0,\\
0,&E_t=0,\ E_v>0,\\
\tfrac12,&\text{otherwise}.
\end{cases}
\label{eq:zero_energy}
\end{equation}
Then apply the same exact budget projection. The third case includes complete absence of gradient responses and responses that never occur jointly. For nonzero but identical paired profiles, Eq.~\ref{eq:response_optimum} already specifies the equal fraction.

\end{document}